\documentclass[sigconf]{acmart}
\usepackage{graphicx}
\usepackage{balance}
\usepackage{booktabs}
\usepackage{caption}
\usepackage{subcaption}
\AtBeginDocument{%
  \providecommand\BibTeX{{%
    \normalfont B\kern-0.5em{\scshape i\kern-0.25em b}\kern-0.8em\TeX}}}

\copyrightyear{2024}
\acmYear{2024}
\setcopyright{rightsretained}
\acmConference[CHI EA '24]{Extended Abstracts of the CHI Conference on
Human Factors in Computing Systems}{May 11--16, 2024}{Honolulu, HI, USA}
\acmBooktitle{Extended Abstracts of the CHI Conference on Human Factors in
Computing Systems (CHI EA '24), May 11--16, 2024, Honolulu, HI, USA}
\acmDOI{10.1145/3613905.3650855}
\acmISBN{979-8-4007-0331-7/24/05}

\begin{document}

\title{Should I Help a Delivery Robot? Cultivating Prosocial Norms through Observations}
\author{Vivienne Bihe Chi}
\email{vivienne_chi@brown.edu}
\orcid{1234-5678-9012}
\affiliation{%
  \institution{Brown University}
  \city{Providence}
  \state{RI}
  \country{USA}
}

\author{Shashank Mehrotra}
\email{shashank\_mehrotra@honda-ri.com}
\orcid{0000-0002-6749-3773}
\affiliation{%
  \institution{Honda Research Institute USA, Inc.}
  \streetaddress{70 Rio Robles}
  \city{San Jose}
  \state{CA}
  \country{USA}}
  
\author{Teruhisa Misu}
\email{tmisu@honda-ri.com}
\orcid{0000-0002-6398-9245}
\affiliation{%
  \institution{Honda Research Institute USA, Inc.}
  \streetaddress{70 Rio Robles}
  \city{San Jose}
  \state{CA}
  \country{USA}}
  
\author{Kumar Akash}
\email{kakash@honda-ri.com}
\orcid{0000-0003-2807-0943}
\affiliation{%
  \institution{Honda Research Institute USA, Inc.}
  \streetaddress{70 Rio Robles}
  \city{San Jose}
  \state{CA}
  \country{USA}}

\renewcommand{\shortauthors}{Chi, et al.}

\renewcommand{\shortauthors}{Chi, et al.}

\begin{abstract}

  We propose leveraging prosocial observations to cultivate new social norms to encourage prosocial behaviors toward delivery robots.
  With an online experiment, we quantitatively assess updates in norm beliefs regarding human-robot prosocial behaviors through observational learning.
  Results demonstrate the initially perceived normativity of helping robots is influenced by familiarity with delivery robots and perceptions of robots' social intelligence.
  Observing human-robot prosocial interactions notably shifts peoples' normative beliefs about prosocial actions; thereby changing their perceived obligations to offer help to delivery robots.
  Additionally, we found that observing robots offering help to humans, rather than receiving help, more significantly increased participants’ feelings of obligation to help robots. 
Our findings provide insights into prosocial design for future mobility systems. Improved familiarity with robot capabilities and portraying them as desirable social partners can help foster wider acceptance. Furthermore, robots need to be designed to exhibit higher levels of interactivity and reciprocal capabilities for prosocial behavior. 



\end{abstract}



\begin{teaserfigure}
    \centering
  \includegraphics[width=.9\textwidth]{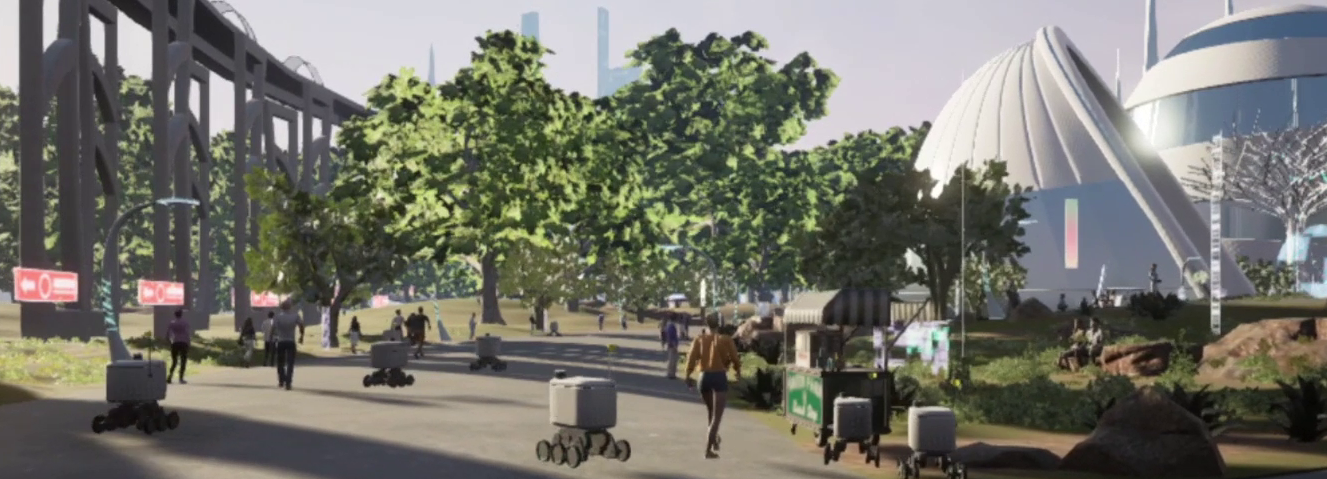}
  \caption{A view of the futuristic environment presented in the online study to the participants. In this city, we show the co-existence of human and AI agents and their prosocial interactions.}
  \Description{A screenshot of the simulation environment, which shows a futuristic urban landscape where approximately equal amounts of  human pedestrians and delivery robots share a road in an urban parkway.}
  \label{fig:simulator}
\end{teaserfigure}

\maketitle
\section{Introduction and Background}
Autonomous mobility systems such as delivery robots are increasingly present in social spaces such as hotels, restaurants, hospitals, and public roads.
Beyond their primary function of transporting items from one location to another, these mobility robots are emerging as active participants in social interactions, both with the intended users and bystanders \cite{Astrid2020,Nomura2016,brscic2015,pelikan_encountering_2024}. Emerging evidence indicates that these robots are perceived by people as entities beyond mere tools, often being anthropomorphized and subjected to social expectations \cite{weinberg_sharing_2023}. This evolving dynamic suggests that mobility robots represent a unique social category, necessitating thoughtful integration into the fabric of human society.

The introduction of mobility robots on roads brings potential benefits. Yet, their widespread adoption is hindered by a lack of transparency and understanding, as well as limited public acceptance compared to traditional mobilities \cite{abrams2021,Wirtz2018,weinberg_sharing_2023}. As highlighted by \cite{valdez_humans_2023,Han2023,Bennett_2021}, the deployment of delivery robots in urban areas has caused noticeable tensions. To fully harness the potential of autonomous mobility and foster a harmonious relationship between human road users and autonomous agents on roads, it's crucial to develop strategies that encourage prosocial interactions between humans and robots \cite{dorrenbacher_towards_2022,sahin_workshop_2021}.


Prosocial behaviors, defined as voluntary behavior intended to benefit others without guaranteed rewards to the helper \cite{oliveira_human-robot_2021,eisenberg_relation_1987,grusec_helping_2011}, are prevalent in mobility contexts. This includes actions like yielding and signaling, where road users commonly offer to and expect to receive assistance from one another \cite{kaye_road_2022,ward_traffic_2020}. Recruiting help from human road users is crucial for addressing the challenges faced by delivery robots deployed in dynamic environments \cite{dobrosovestnova_little_2022}.


Prior research in Human-Computer Interaction (HCI) has demonstrated success in eliciting human help in certain controlled and real-life environments \cite{srinivasan_help_2016,weiss_robots_2010,paiva_engineering_2018,maeda_influencing_2021,peter_can_2021,potinteu_insight_2023,Hang2022}, with decisions to help influenced by factors such as the situational context, the robot's physical design (ranging from anthropomorphic to purely functional) \cite{kim2014}, the affective response elicited by the robot (such as the psychology of `kawaii', \cite{nittono_psychology_2022}), and the robot's signaling for help \cite{boos2022,MARTIN2020,SANTOS2011}. 

However, the observed prosocial behaviors towards robots in one-off interactions may still be influenced by the novelty effect \cite{dobrosovestnova_little_2022}. The reliability of situation- and robot-related factors in consistently eliciting prosocial norms across people who may have different perceptions around interactions with robots remains questionable \cite{bajones_investigating_2017}. In response, we propose to mobilize social observation \cite{brohmer_inspired_2019} and conformity to instigate new human-robot intergroup prosocial norms. Social norms, which monitor, punish, and reward human actions \cite{Fehr2004,bicchieri_grammar_2006}, are pivotal in promoting prosocial behavior \cite{gachter2013,batson_four_2011} \cite{andersson_norms_2022,borinca_ingroup_2022} and can lead to enduring behavioral changes \cite{gavrilets_collective_2017,prentice_engineering_2020,Bicchieri2017}.

This paper reports on a randomized controlled experiment in a high visual-fidelity simulation environment to investigate how observational learning can promote prosocial behavior towards delivery robots. We examine two types of observations---humans helping robots and robots helping humans---to compare their effectiveness in instigating these prosocial norms. 

This paper makes several key contributions to the field. Firstly, it demonstrates the feasibility of using social observation and conformity to establish new prosocial norms between humans and robots. Secondly, it identifies the impact of different prosocial observation scenarios, specifically comparing instances where a robot acts as a helper versus scenarios where the robot is being helped, on the development of prosocial norm beliefs. Thirdly, this research adds to the literature on human-delivery robot interactions by introducing a quantitative, randomized controlled experiment anchored in psychological theories of social norms. Lastly, it provides insights for the interaction design of future mobility robots, highlighting the significance of the prosocial behaviors displayed by these robots in promoting prosocial norms in their interactions.

\section{Research Objectives}
\label{RQs}
As a first step towards exploring the use of social observation as a tool to foster new prosocial norms in human-robot interactions, we conducted an online experiment on a high visual-fidelity simulation platform to answer the following research questions (RQs):
    \begin{itemize}
        \item  RQ1: What are the prevailing normative beliefs about prosocial interactions with robots?
        \item RQ2: What factors influence people's perceptions of the normativity of assisting robots?
        \item RQ3: Can observations lead to a change in beliefs about prosocial norms, and how do these changes impact the perceived obligation to act prosocially?
        \item RQ4: Which observation type-- robots acting as helpers or as beneficiaries-- more effectively fosters prosocial behavioral norms towards robots?
    \end{itemize}
\section{Methods}
    \subsection{Study Design}
     This study utilized a mixed design, incorporating both between-subject and within-subject factors. Specifically, it featured three between-subject observation conditions: 1) human helping human, 2) human helping robot, and 3) robot helping human. The within-subject factor involved three scenarios: 1) warning of oncoming car, 2) notification of road closure, and 3) picking up misplaced trash. Each participant was exposed to all three scenarios. Participants were randomly assigned to one of the observation conditions and were subjected to repeated observations within their assigned condition.
       
    \subsection{Participants}
        The online study sample comprised 210 native English speakers in the United States, all recruited on \href{http://prolific.co}{Prolific}; $47.1\%$ were female-identifying.
        Participants' ages ranged from 19 to 75, with a median of 38.5. $31.9\%$ of the participants live in urban areas, $53.8\%$ in suburbs and $12.9\%$ in rural areas. 
        The mean rating of participants' self-reported familiarity with robots was 2.8 on a scale from 1 to 7. Research protocols and procedures were approved by the bioethics committee (Anonymized).
        
    \subsection{Materials}
    The online study was conducted using video recordings of a custom high visual-fidelity simulator. The simulator environment is rendered using Unreal Engine 5.1.1 \cite{UnrealEngine}. The environment represented an urban city with the diffusion of multiple road users sharing the spaces with no dedicated space for road users. The virtual environment aims to recreate smart cities planned and encouraged in several cities across the EU and North America \cite{cugurullo2021transition}. 
    To this effect, the participants are shown to be walking in a park-like environment with different types of road actors, including pedestrians (of all age groups and genders), food cart vendors, delivery robots, and small cart-like self-driving cars. The simulator environment is shown in Figure \ref{fig:simulator}.

\begin{figure}[ht]
   \begin{subfigure}[b]{0.236\textwidth}
         \centering
         \includegraphics[width=\textwidth]
         {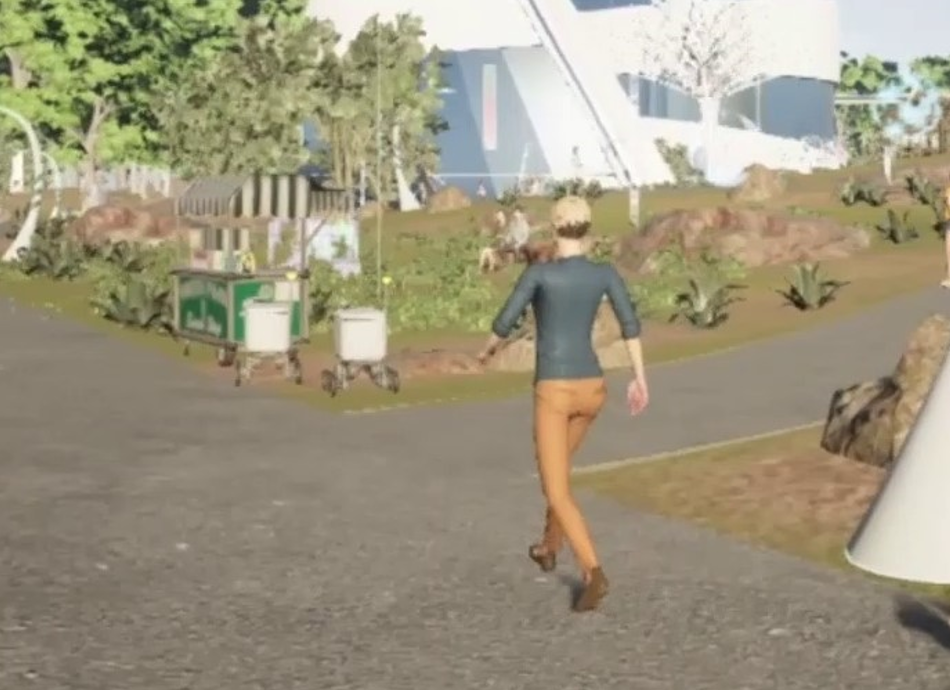}
         \caption{}
   \end{subfigure}
   \begin{subfigure}[b]{0.236\textwidth}
         \centering
         \includegraphics[width=\textwidth]
         {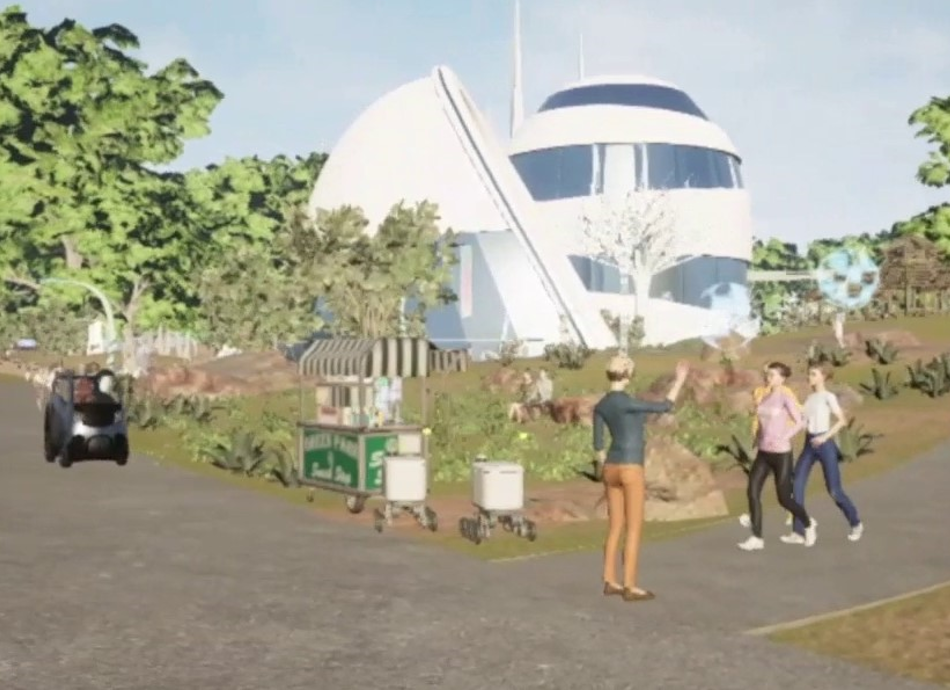}
         \caption{}
   \end{subfigure}
   \begin{subfigure}[b]{0.236\textwidth}
         \centering
         \includegraphics[width=\textwidth]
         {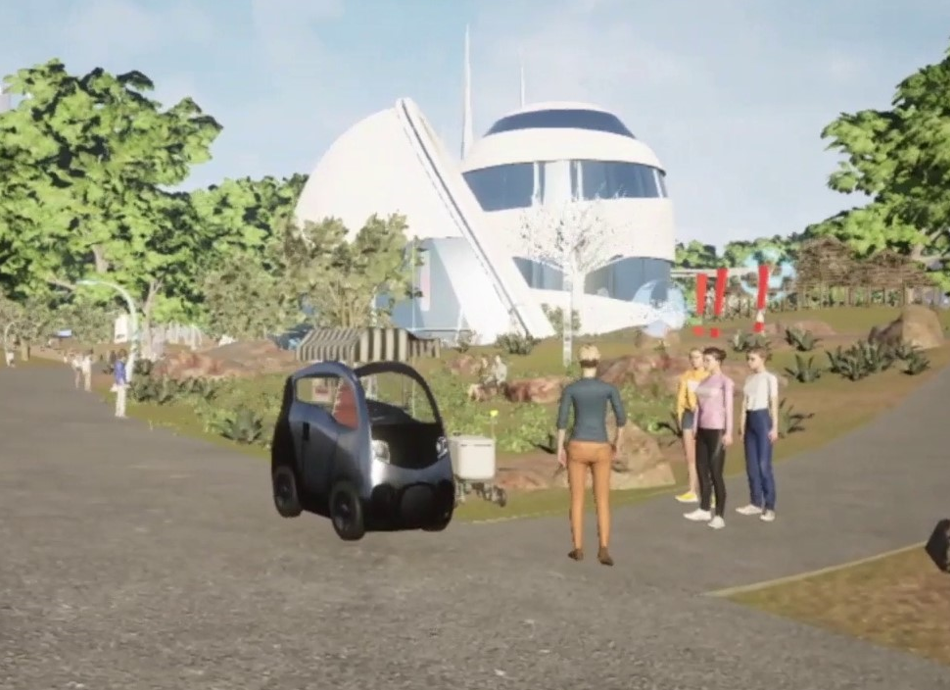}
         \caption{}
   \end{subfigure}
   \begin{subfigure}[b]{0.236\textwidth}
         \centering
         \includegraphics[width=\textwidth]
         {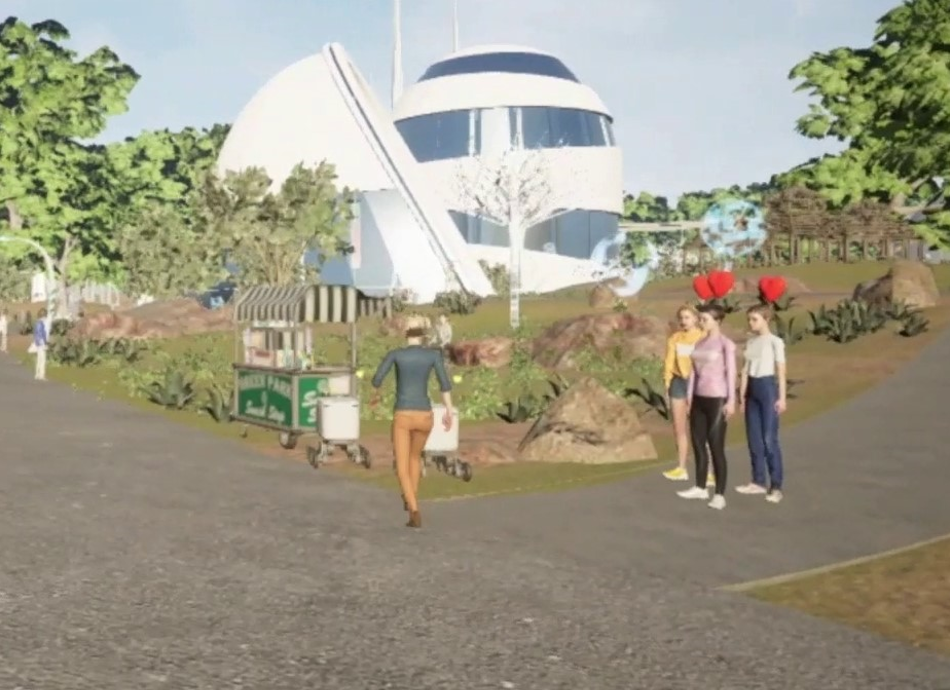}
         \caption{}
   \end{subfigure}
    \caption{Snippets of Scenario 1: warning of oncoming car. (a) Pedestrians walk on a park-like road. (b) Pedestrian warns a group of pedestrians about a fast-approaching car. (c) The group of pedestrians stops and lets the car go past. (D) The group expresses gratitude towards the pedestrians' prosocial behavior}
            \Description{ Four screenshots of the simulation environment, showing the prosocial scenario 1--  warning of oncoming car. (a) Pedestrians walk on a park-like road. (b) Pedestrian warns a group of pedestrians about a fast-approaching car. (c) The group of pedestrians stops and lets the car go past. (D) The group expresses gratitude towards the pedestrians' prosocial behavior}
            \label{fig:scenario1}
\end{figure}

            
        To assess people's attitudes of prosocial behaviors toward other road users, we placed participants into specific scenarios. Reflecting on the challenges discussed at the beginning of this paper, we focused on designing scenarios in a future mobility context where delivery robots are widely present and share roads with pedestrians. We specifically targeted prosocial behaviors that 1) would incur a small cost to the helper, 2) are not already deemed as a requirement, allowing room for learning through observation, and 3) are within the capability of a delivery robot to reciprocate.
        
        Through a collaborative iterative design process, we crafted three prosocial interaction scenarios-- 1) warning of oncoming car, 2) notification of road closure, and 3) picking up misplaced trash.
        Each scenario involves a `helper' (the actor performing the prosocial behavior) and a `beneficiary' (the recipient of the help), who could be either a human pedestrian or a delivery robot.
        Videos of the pedestrian walking in the simulation environment were recorded from a third-person view for observation trials (where participants watch the scenarios as bystanders) and a first-person view for decision trials (where participants assume the role of a potential helper). The videos also included visual cues to aid participants' understanding of the scenarios. Figure \ref{fig:scenario1} illustrates a sample scenario of warning of oncoming cars. Full videos of scenarios are included in the supplementary materials.

    \subsection{Measures}
    At the outset of the study, to gauge people's prior experience and understanding of delivery robots prior to any exposure to experimental stimuli or manipulations,  we measured people's self-reported robot familiarity (\textit{"How familiar are you with delivery robots?"} measured on a 7-point scale from \textit{Not at all} to \textit{Very familiar}) and their perceived social intelligence of delivery robots using the short form Perceived Social Intelligence (PSI) Scale \cite{barchardperceived}. People's perceived social intelligence in delivery robots was measured along two dimensions---social presentation (a robot's appeal as a social partner) and social information processing (a robot's capabilities to work alongside humans). 
    Furthermore, we employed the Prosociality Scale \cite{gian_vittorio_caprara_misura_2005} to measure participants' inherent tendencies towards prosocial behavior, as it is anticipated to be a predictive factor for prosocial norm beliefs.
           
    In the main experiment, we first measured participants' baseline normative beliefs regarding the expected prosocial actions of either a human pedestrian or a delivery robot in the three mobility scenarios.  This was done by presenting participants with normative statements such as ``A human pedestrian [observed helper] is \rule{0.5cm}{0.1mm} to inform a delivery robot [observed beneficiary] of the road closure.'' Participants expressed their perceived degree of normativity by moving a slider bar on a scale of -10 (Prohibited), -5 (Discouraged), 0 (Allowed), 5 (Encouraged), 10 (Required). Following two rounds of observation of their assigned condition, we measured participants' updated normative beliefs. 
        
    In the decision trials that followed, participants were presented with normative statements such as ``I am \rule{0.5cm}{0.1mm} to inform a delivery robot [potential beneficiary] of the road closure.'' They used the same slider bar and scale to indicate their sense of obligation to perform prosocial actions in each scenario.

    \subsection{Procedure}
    Upon obtaining informed consent, we administered a pre-experiment survey to gather demographic data, assess their self-reported familiarity with delivery robots, and evaluate their current perception using the 20-item PSI scale \cite{barchardperceived}.

    Participants were then introduced to the experimental context through a brief text. This text aimed to immerse them in a hypothetical urban environment set in the year 2050, characterized by the widespread use of autonomously operated delivery robots that navigate alongside human pedestrians.

   Subsequently, they viewed a one-minute narrated introduction video, designed to be representative of the series of video stimuli they would encounter throughout the experiment.  
   The videos depicted a futuristic city center from a first-person perspective, highlighting shared street scenes with equal numbers of human pedestrians and mobility robots. 
   Participants were instructed to view these videos as if they were experiencing the scene through Augmented Reality glasses while walking through the city center. The videos are included in the supplementary materials.
       
    The main experiment commenced with an initial series of observation trials. Midway through these videos, right before the prosocial acts took place, we assessed participants' baseline normative beliefs for each of the three scenarios.
    This was followed by the second round of observation trials, featuring the same scenarios in a varied sequence. After these repeated observations, we measured the participants' updated normative beliefs.

    Subsequently, participants engaged in decision trials. In these trials, the videos depicted similar scenarios, but occurring in closer proximity to the participant's ego, thereby prompting them to gauge their own inclination to act prosocially. Participants rated the extent to which they felt normatively compelled to take action.

    Finally, at the end of the main experiment, we evaluated participants' prosocial inclinations using a survey developed by \cite{gian_vittorio_caprara_misura_2005} and solicited general feedback for the study through a post-experiment survey.
    
\section{Results}
 In this section, we present our analyses of participants' normative beliefs measured at three distinct time points: baseline normative beliefs prior to prosocial observations, post-observation normative beliefs, and the perceived normativity of helping humans and robots during decision trials. Descriptive statistics are summarized in Table \ref{tab:summary}.
 
\begin{table*}[ht]
 \caption{Descriptive statistics of norm beliefs measured for the three observation conditions at various time points on a scale from -10 (Prohibited) to 10 (Required)}
 \label{tab:summary}
\resizebox{\textwidth}{!}{%
\begin{tabular}{@{}lllll@{}}
\toprule
Observation Condition    & \multicolumn{1}{c}{\begin{tabular}[c]{@{}c@{}}Baseline \\ normative belief\end{tabular}} & \multicolumn{1}{c}{\begin{tabular}[c]{@{}c@{}}Post-observation \\ normative belief\end{tabular}} & \multicolumn{1}{c}{\begin{tabular}[c]{@{}c@{}}Decision trial \\ rormative rating,\\ Robot beneficiary\end{tabular}} & \multicolumn{1}{c}{\begin{tabular}[c]{@{}c@{}}Decision trial \\ normative rating,\\ Human beneficiary\end{tabular}} \\ \midrule
robot helping human (RH) & M = 2.58, SD = 3.65                                                                      & M = 4.18, SD = 3.09                                                                              & M = 4.31, SD = 3.18                                                                                                 & M = 4.51, SD = 3.44                                                                                                 \\
human helping robot (HR) & M = 3.19, SD = 4.69                                                                      & M = 4.72, SD = 3.13                                                                              & M = 2.23, SD = 3.56                                                                                                 & M = 4.58, SD = 3.41                                                                                                 \\
human helping human (HH) & M = 4.01, SD = 4.96                                                                      & M = 6.99, SD = 3.47                                                                              & M = 1.58, SD = 4.42                                                                                                 & M = 4.49, SD = 3.64                                                                                                 \\ \bottomrule
\end{tabular}%
}
\end{table*}

\subsection{Baseline normative belief}
\label{baseline_belief}
    First, we examined participants' initial normative beliefs about helping human pedestrians versus delivery robots in realistic mobility contexts (\textbf{RQ1}).
    To do this, we used a mixed-effects model with participant and scenario as random effects, and observation condition (robot-helping-human, human-helping-robot, or human-helping-human, see Table \ref{tab:summary}), participant prosociality (measured with the Prosociality Scale by \cite{gian_vittorio_caprara_misura_2005}), and their interactions as fixed effects. Reverse Helmert contrasts were used to compare the intergroup helping conditions [robot-helping-human(RH) vs. human-helping-robot(HR)] against the within-group helping condition [human-helping-human(HH)].
    
    Before any experimental exposure, intergroup helping [human-helping-robot(HR) or robot-helping-human(RH)] was perceived as less normative ($t = -2.41, p =.01$) compared to within-group helping [human-helping-human(HH)]. However, no significant baseline differences were observed between the two treatment groups ($t = 1.57, p = .1$). As expected, participants' prosocial inclinations positively correlated with higher normative ratings of prosocial behaviors ($t = 2.21, p = .02$). Those with lower prosocial inclinations viewed intergroup (Human-Robot) helping as less normative compared to within-group (Human-Human) helping ($t = -1.90, p = .05$).

    However, a participant's general prosociality was not the strongest predictor of their beliefs about the normativity of humans helping delivery robots. To address \textbf{RQ2}, we modified the mixed-effects model to include factors like participants' familiarity with delivery robots and their perceptions of the robots' social information processing abilities and social presentation characteristics along with the interactions to predict the baseline normative belief of humans helping robots.
    The analysis revealed that the perceived social presentation characteristics ($t = 2.66, p = .01$) and social information processing abilities ($t = -3.04, p <.005$) had a greater impact than prosociality ($t = 1.09, p = .28$) in shaping these beliefs. Higher ratings of robots' social presentation traits correlated with stronger normative beliefs in favor of intergroup prosocial interactions. Furthermore, perceived social information processing abilities interacted with familiarity. Lower perceived abilities led to stronger normative beliefs in favor of helping robots, especially among participants less familiar with delivery robots ($t= 2.93, p <.005$).
    
    Further analysis showed that the perceived social intelligence of robots mediates the effect of familiarity on the normative belief of helping robots ($ACME= .19, p<.001$, $ADE= .27, p =.14$). 
    In essence, greater familiarity with delivery robots led to higher perceptions of their social intelligence, which in turn reduced the perceived obligation to assist them.


\subsection{Post-observation normative belief change}
\label{normbeliefchange}
    After completing the two rounds of prosocial observation trials, we assessed changes in participants' normative beliefs. A repeated measures ANOVA was conducted to assess these changes, confirming a significant update ($F(1) = 4.06, p < .05$) in participants' norm beliefs between the two measurement points (before and after the observations), indicating the effectiveness of the observations in altering people's prosocial norm beliefs (\textbf{RQ3}).
    A pair of simple-effect analyses for the two intergroup prosocial observation conditions [robot-helping-human(RH) vs. human-helping-robot(HR)] revealed that changes in the belief that humans should assist robots (HR) were influenced by participants' prior familiarity with delivery robots ($t = 2.65, p = .01$), their perceived social intelligence of these robots (measured by the PSI scale \cite{barchardperceived}, $t = 2.61, p = .01$), 
     Additionally, there was a significant interaction between these two factors ($t = -2.9, p < .005$), indicating that the effect of familiarity on norm belief change was moderated by the level of perceived social intelligence in the robots. 
    In contrast, these factors did not affect changes in norm beliefs regarding whether delivery robots should assist human pedestrians (RH, $t_s < .65, p_s > .5$).
    

\subsection{Decision trial normative rating}
\label{decision_trials}
\begin{figure}[ht]
    \centering
    \includegraphics[width=\linewidth]{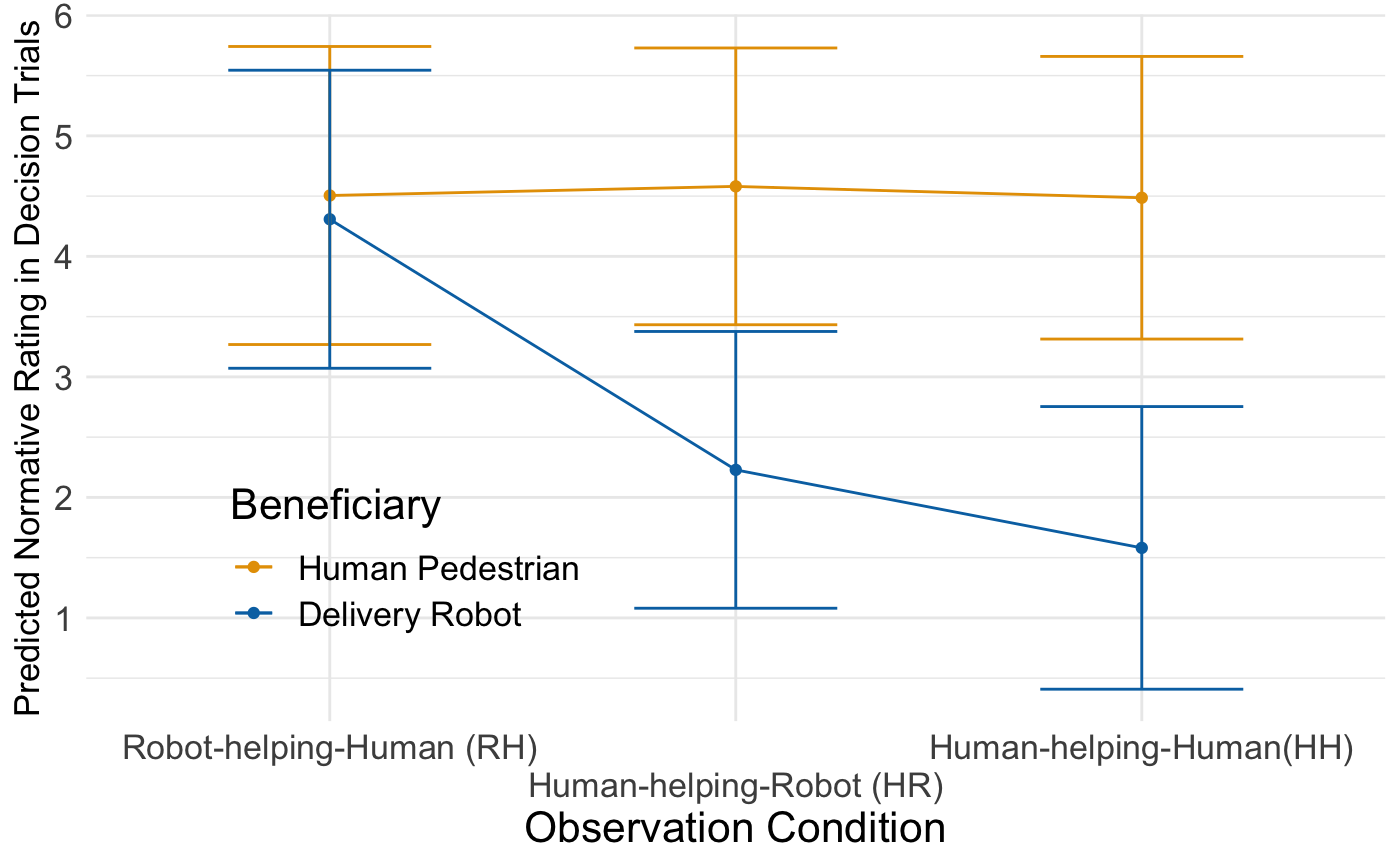}
    \caption{{ Distribution of the normative behavior rating across the three observation conditions}}
    \Description{Line graph showing the predicted normative ratings in decision trials for the two beneficiary types-- human pedestrian or delivery robot across the three observation conditions of robot-helping-human(RH), human-helping-robot(HR) and human-helping-human(HH). The ratings for human pedestrian beneficiaries remain consistently high for all three observation conditions. The rating for delivery robot beneficiaries is comparable to that for human pedestrian beneficiaries only in the robot-helping-human condition. In the other two conditions, the rating for delivery robot beneficiaries is significantly lower than for human pedestrian beneficiaries.}
    \label{fig:intention_act}
\end{figure}

   Finally, we wanted to understand the impact of previous observations on individuals' perceived obligation to act prosocially in similar situations. 
    We built a mixed-effects model predicting people's normative rating in decision trials (i.e., their own sense of obligation to assist a delivery robot in a given situation), with participant and scenario as random effects and fixed effects of the observation condition (see Table\ref{tab:summary}), beneficiary (potential recipient of help), and their interaction. 
    The results displayed a general trend that participants across all observation conditions felt a stronger obligation to assist human pedestrians compared to robots ($t = 11.17 , p <.001$). 
    Notably, observations of robots helping humans (RH) were significantly more influential ($t = 5.33, p < .001$) in fostering a sense of obligation to help delivery robots than observations of humans helping robots (HR). This finding, illustrated in Figure \ref{fig:intention_act}, decisively answers our fourth research question (\textbf{RQ4}), demonstrating that observations of robots acting as helpers, rather than beneficiaries, more effectively promote prosocial behavioral norms towards robots.
    
    To examine the impact of changes in normative beliefs (described in section \ref{normbeliefchange}) on participants' expressed obligation to assist robots during decision trials (addressing the second part of \textbf{RQ3}), we extended the previous model to incorporate post-observation norm belief changes as a predictive factor.
    Findings revealed a notable distinction between the two treatment observation groups (RH and HR, outlined in Table \ref{tab:summary}). Specifically, positive changes in the belief that robots should help humans (RH) significantly increased participants' feelings of obligation to help robots ($t = 2.79, p =.005$). 
    This suggests that participants who observed robots helping humans not only learned but also began to internalize the robot-helping-human norm, leading to a heightened sense of reciprocal obligation towards robots.

    Contrary to what might be expected, observing humans helping robots (HR) did not yield a similar effect. This outcome challenges the assumption that direct observation of human-helping-robot norms would more directly influence learning. Learning about human-helping-robot norms from a third-person viewpoint may not be as compelling for norm internalization, particularly in the absence of norm enforcement mechanisms. Our interpretation highlights the role of reciprocal expectations, rooted in the robot-helping-human norm, in fostering a self-motivated drive for prosocial behavior.

\section{Discussion and Conclusion}
  
  Drawing on psychological theories of social norms, we proposed leveraging prosocial observations to cultivate new prosocial norms toward delivery robots. Our randomized controlled online experiment quantitatively evaluated changes in perceived normativity related to human-robot prosocial behaviors at three stages: baseline (Section \ref{baseline_belief}) and post-observations trials (Section \ref{normbeliefchange}, and during the subsequent decision trials (Section \ref{decision_trials}) where participants assume the role of potential helpers. 

  Study results address the four research questions outlined in Section \ref{RQs}. Firstly, addressing RQ1 and RQ2, we found that people's initial norm beliefs of helping robots are influenced by individuals' familiarity with delivery robots and their perceptions of these robots' social intelligence. This suggests that educating community members about mobility robot capabilities to improve familiarity and portraying them as desirable social partners can enhance the acceptance of mobility robots in public spaces. 
Next, in response to RQ3, our results indicate that the observations notably shift normative beliefs about prosocial actions, and subsequently influenced people's perceived obligations to offer help to delivery robots. This illustrates the effectiveness of leveraging observational learning to induce norm belief changes.
 Lastly, addressing RQ4, our experiment, which assigned participants to one of three observation conditions (Robot-helping-Human, Human-helping-Robot, Human-helping-Human), revealed that observing robots assisting humans (rather than being assisted) more significantly increased participants’ feelings of obligation to help robots. Our interpretation of this result highlights the role of reciprocal expectations in human-robot interactions. To encourage prosocial human behavior towards robots in real-world settings, it is crucial to design robots that exhibit higher levels of interactivity and the ability to reciprocate assistance.

    
The presented study is subject to several limitations.
First, the study was conducted online, presenting scenarios through videos and relying exclusively on self-reported measures in response to these stimuli. Such an approach, while accessible and broad in reach, may not fully capture the complexity of real-world interactions or accurately predict actual behavior toward robots.
   Secondly, by situating the study in a futuristic context, we aimed to shift focus from the safety and performance of delivery robots to the possibility of engaging with them socially. However, it remains uncertain if these findings can instigate real-life behavioral changes or if such changes would persist beyond the experimental session.
   Finally, our research utilized a single generic model of a delivery robot, leaving the applicability of our results to other robot types within and beyond the mobility context unexplored. 
   To mitigate some of these limitations, we plan to conduct an in-person study using virtual reality. This method will enhance realism and immersion, allowing for direct measurement of prosocial behaviors via eye tracking, motor responses, and physiological data. Furthermore, incorporating a qualitative study will deepen our understanding of prosociality through triangulated measures, providing a richer analysis of human-robot interactions.
    
    Overall, our research contributes to the field by identifying the key factors of individual prosociality inclinations, familiarity with a specific type of robot, and the perceived social intelligence of a robot in shaping the prevailing normative beliefs in human-robot prosocial interactions. We demonstrate how observational learning from robot-to-human prosocial interactions can promote human prosocial behaviors towards delivery robots, fostering new norms that enhance the acceptance and integration of mobility robots in society, thereby advancing the harmonious coexistence of humans and mobility robots in public spaces.
    

\bibliographystyle{ACM-Reference-Format}
\bibliography{CHI24LBW}

\appendix

\end{document}